\documentclass[runningheads]{llncs}

\usepackage{booktabs} 
\usepackage{array}
\usepackage{multirow}

\usepackage{graphicx}
\usepackage{hyperref}

\usepackage[dvipsnames]{xcolor}

\begin{document}
\title{Digital Editions as Distant Supervision \\ for Layout Analysis of Printed Books\thanks{Authors contributed equally. We would like to thank the Deutsches Textarchiv, the Women Writers Project, and Rui Dong for collecting data and running the baseline OCR system. This work was supported in part by a National Endowment for the Humanities Digital Humanities Advancement Grant (HAA-263837-19) and the Andrew W. Mellon Foundation’s Scholarly Communications and Information Technology program. Any views, findings, conclusions, or recommendations expressed do not necessarily reflect those of the NEH or Mellon.}}
\titlerunning{Digital Editions as Distant Supervision}
%

\author{Alejandro H. Toselli\inst{1} \and
Si Wu\inst{1} \and
David A. Smith\inst{1}}


\institute{Khoury College of Computer Sciences\\
Northeastern University, Boston, MA 02115, U.S.A. \\
\email{\{\a.toselli,wu.si2,davi.smith\}@northeastern.edu}}

\maketitle              
\begin{abstract}

Archivists, textual scholars, and historians often produce digital editions of historical documents.  Using markup schemes such as those of the Text Encoding Initiative and EpiDoc, these digital editions often record documents' semantic regions (such as notes and figures) and physical features (such as page and line breaks) as well as transcribing their textual content.  We describe methods for exploiting this semantic markup as distant supervision for training and evaluating layout analysis models.  In experiments with several model architectures on the half-million pages of the \emph{Deutsches Textarchiv} (DTA), we find a high correlation of these region-level evaluation methods with pixel-level and word-level metrics.  We discuss the possibilities for improving accuracy with self-training and the ability of models trained on the DTA to generalize to other historical printed books.

\end{abstract}
\section{Introduction}
\label{sec:intro}

Expanding annotated data for training and evaluation has driven progress in automatic layout analysis of page images.  Most commonly, these annotated datasets are produced by manual annotation or by aligning the input documents with the typesetting information in PDF and similar formats \cite{zhong19:icdar}.

This paper describes methods for exploiting a further source of information for training and testing layout analysis systems: \textbf{digital editions with semantic markup}.  Many researchers in archival and literary studies, book history, and digital humanities have focused on digitally encoding books from the early modern period (from 1450) and the nineteenth century \cite{franzini12:_catal_digit_edition}. These editions have often employed semantic markup---now usually expressed in XML---to record logical components of a document, such as notes and figures, as well as physical features, such as page and line breaks.

Common markup schemes---as codified by the Text Encoding Initiative, EpiDoc, or others---have been mostly used for ``representing those features of textual resources which need to be identified explicitly in order to facilitate processing by computer programs'' \cite[p.~xvi]{TEIp5}.  Due to their intended uses in literary and linguistic analysis, many digital editions abstract away precise appearance information.  The typefaces used to distinguish footnotes from body text, for example, and the presence of separators such as horizontal rules or whitespace, often go unrecorded in digital editions, even when the semantic separation of these two page regions is encoded.

After discussing related work on modeling layout analysis (\S\ref{sec:related}), we describe the steps in our procedure for exploiting digital editions with semantic markup to produce annotated data for layout analysis.\footnote{For data and models, see \url{https://github.com/NULabTMN/PrintedBookLayout}}

First (\S\ref{sec:markup}), we analyze the markup in a corpus of digital editions for those elements corresponding to page-layout features.  We demonstrate this analysis on the \emph{Deutsches Textarchiv} (DTA) in German and the \emph{Women Writers Online} (WWO) and \emph{Text Creation Partnership} (TCP) in English.
    
Then (\S\ref{sec:align}), we perform forced alignment to link these digital editions to page images and to link regions to subareas on those page images.  For the DTA, which forms our primary test case, open-license images are already linked to the XML at the page level; for the WWO, we demonstrate large-scale alignment techniques for finding digital page images for a subset of books in the Internet Archive.  For pages with adequate baseline OCR, we also align OCR output with associated page coordinates with text in regions in the ground-truth XML.  Some page regions, such as figures, are not adequately analyzed by baseline OCR, so we describe models to locate them on the page.

In experimental evaluations (\S\ref{sec:experiments}), we compare several model architectures, pretrained, fine-tuned, and trained from scratch on these bootstrapped page annotations.  We compare region-level detection metrics, which can be computed on a whole semantically annotated corpus, to pixel- and word-level metrics and find a high correlation among them.

\section{Related Work}
\label{sec:related}

Perhaps the largest dataset proposed recently for document layout analysis is PubLayNet \cite{zhong19:icdar}. The dataset is obtained by matching XML representations and PDF articles of over 1 million publicly available academic papers on PubMed Central\textsuperscript{TM}.  This dataset is then used to train both Faster-RCNN and Mask-RCNN to detect text, title, list, table, and figure elements. Both models use ResNeXt-101-64x4d from Detectron as their backbone. Their Faster-RCNN and Mask-RCNN achieve macro mean average precision (MAP) at intersection over union (IOU) [0.50:0.95] of 0.900 and 0.907 respectively on the test set.

Newspaper Navigator \cite{lee2020newspaper} comprises a dataset and model for detecting non-textual elements in the historic newspapers in the \emph{Chronicling America} corpus. The model is a finetuned R50-FPN Faster-RCNN from Detectron2 and is trained to detect photographs, illustrations, maps, comics/cartoons, editorial cartoons, headlines, and advertisements. The authors report a MAP of 63.4\%.


U-net was first proposed for medical image segmentation \cite{Ronneberger15:springer}. Its architecture, based on convolutional layers, consists of a down-sampling analysis path (encoder) and an up-sampling synthesis path (decoder) which, unlike regular encoder-decoders, are not decoupled. There are skip connections to transfer fine-grained information from the low-level layers of the analysis path to the high-level layers of the synthesis path as this information is required to accurately generate fine-grained reconstructions.  In this work, we employ the U-net implementation \emph{P2PaLa}\footnote{\url{https://github.com/lquirosd/P2PaLA}} described in~\cite{quiros18:arXiv} for detection and semantic classification of both text regions and lines. This implementation has been trained
and tested on different publicly available datasets:
cBAD\,\cite{cBAD19} for baseline detection, and
Bozen\,\cite{Sanchez16} and OHG\,\cite{Quiros18} for both text region
classification and baseline detection. Reported mean intersection over
union results are above 84\% for region and baseline detection on the
Bozen dataset.
It is worth noting that the U-net implementation is
provided with a weighted loss function
mechanism~\cite{Paszke2016ENetAD}, which can mitigate possible
class imbalance problems.

Kraken, an OCR system forked from \emph{Ocropy}, uses neural
networks to perform both document layout analysis and text
recognition.%
\footnote{See \url{http://kraken.re} and \url{https://github.com/ocropus/ocropy}.}
For pixel classification in layout analysis, Kraken's network
architecture was designed for fewer memory resources than U-net. Roughly, it comprises down-sampling convolutional layers
with an increasing number of feature maps followed by BLSTM blocks for
processing such feature maps in both horizontal and vertical
directions\,\cite{Kiessling20:icfhr}. The final convolutional layer,
with sigmoid activation function, outputs probability maps of regions
and text lines.
Kraken's model for baseline detection has been trained and tested on
the public dataset BADAM\,\cite{Kiessling19:BADAM} and also on the
same datasets as P2PaLA. For region detection, Kraken obtained mean
intersection over union figures are 0.81 and 0.49 for Bozen and OHG
datasets respectively.

Several evaluation metrics have been commonly employed for document layout analysis.  The Jaccard Index, also known as intersection over union (iu), is one of the most popular pixel-level evaluation measures used in ICDAR's
organized competitions related with document layout analysis as
\cite{Gao17:icdar},
\cite{Burie15:icdar}. Likewise this measure has also served as a way
to consider when there is a match between detected objects and their
references as in
\cite{Shi17:icdar}.


\section{Analyzing Ground Truth Markup}
\label{sec:markup}

Scholars of early printed books create digital editions for several purposes, from enabling full-text search to studying language change and literary style to studying the work practices of letterpress compositors.

In this paper, we focus on the ``top-level'' units of layout, which for brevity we call \textbf{regions}.  Within each region, one or more lines follow each other in a sequential reading order (e.g., top-to-bottom or right-to-left).  \emph{Among} regions on a page, no such total order constraint necessarily holds.  Page numbers and running titles, for instance, whether at the top or bottom or a page, do not logically ``precede'' or ``follow'' the main body or footnote text.

We analyze the conventions of encoding these top-level regions in three broad-coverage corpora of historical printed books. The \emph{Deutsches Textarchiv} (DTA) \cite{dta} comprises transcriptions of 1406 German books in XML following Text Encoding Initiative (TEI) \cite{TEIp5} conventions along with over 500,000 page images. The \emph{Women Writers Online} (WWO) \cite{wwo} corpus contains, as of this writing, 336 books of womens' writing in English, transcribed in TEI XML but with no page images of the editions transcribed. In \S\ref{sec:align} below, we discuss a forced alignment process to link a subset of the WWO gold-standard XML to page images from the Internet Archive. The \emph{Text Creation Partnership} (TCP) \cite{tcp} contains TEI XML transcriptions of 32,853 books from the proprietary microfilm series \emph{Early English Books Online}, \emph{Eighteenth-Century Collections Online}, and \emph{Evans Early American Imprints}.

Table~\ref{tab:zones} summarizes the XML encoding conventions used for eight top-level regions in these three corpora.  For precision, we use XPath notation \cite{Dyck:17:XPL}.  All three corpora include some top-level regions such as \textbf{body}, \textbf{figure}, and \textbf{note}.  The source texts that were transcribed to compile a corpus may still of course contain regions not reflected in the XML edition: for example, \textbf{running titles} are present in books from all three corpora, but the DTA is the only corpus that transcribes then.

\begin{table}
\caption{Summary of page zone markup in TEI editions from the \emph{Deutsches Textarchiv} (DTA), \emph{Text Creation Partnership} (TCP), and \emph{Women Writers Online} (WWO).  We remove trailing \texttt{text()} functions from the XPath selectors for simplicity.}\label{tab:zones}

\begin{tabular}{|l|l|l|l|}
\hline
\bf Corpus & \bf Caption & \bf Catchword \\
\hline
DTA & \verb|//figure/*| & \verb|//fw[@type='catch']| \\
TCP & \verb|//figure/*[not(self::figDesc)]| & --- \\
WWO & \verb|//figure/*[not(self::figDesc)]| & \verb|//mw[@type='catch']| \\
\hline
 & \bf Column head & \bf Figure \\ \hline
DTA & \verb|//cb[substring(@n,1,1)!='[']/@n| & \verb|//figure| \\
TCP & --- & \verb|//figure| \\
WWO & --- & \verb|//figure| \\
\hline
 & \bf Note &  \bf Pagination \\ \hline
DTA & \verb|//note| & \verb|//pb[substring(@n,1,1)!='[']/@n| \\
TCP & \verb|//note| & --- \\
WWO & \verb|//notes/note| & \verb|//mw[@type='pageNum']| \\
\hline
 & \bf Running title & \bf Signature \\ \hline
DTA & \verb|//fw[@type='head']| & \verb|//fw[@type='sig']| \\
TCP & --- & --- \\
WWO & --- & \verb|//mw[@type='sig']| \\
\hline
\end{tabular}
\end{table}

Many small elements from the skeleton of the printing forme and other marginal matter are present in early modern books \cite{werner19}.  Both the DTA and WWO record printed \textbf{signature} marks and \textbf{catchwords} at the bottom of the page, which aided printers in assembling printed sheets into books.  The DTA and WWO both transcribe printed \textbf{page numbers}.  The DTA also encodes inferred page numbers, e.g., when the pagination is not explicitly printed on the first page of a chapter, by enclosing the number in square brackets. The DTA transcribes \textbf{running titles} at the top of the page and the \textbf{column heads} that appear in some multi-column page layouts (e.g., in dictionaries).  The TCP does not record any of these minor elements.

All three corpora transcribe \textbf{notes}.  The DTA and TCP insert \texttt{<note>} elements near the reference mark printed in the body text for footnotes and endnotes.  They also transcribe marginal notes inline.  The WWO transcribes all notes in a separate \texttt{<notes>} section at the beginning of each text and links the child \texttt{<note>} elements to references in the body with XML IDREFs.  The text of the notes in the WWO must therefore be associated with the appropriate page record.  In all three corpora, some foot- and endnotes continue onto the next page.  We therefore assign each part of the text of these run-on notes with the appropriate page.

We define the \textbf{body} text as almost all contents of the \texttt{<text>} element that are not described one of the floating or extraneous elements described above and summarized in Table~\ref{tab:zones}.  The few exceptions to this definition in the three corpora we examine are elements recording editorial interventions in the text: the \texttt{<corr>} element in the WWO for corrected spelling and the \texttt{<gap>} element in the TCP for recording gaps in the transcription due to unreadable microfilm images.  The body text is broken into different zones by page breaks (\texttt{<pb>}) and column breaks (\texttt{<cb>}).  The DTA and WWO record line breaks in the editions they transcribe with \texttt{<lb>} milestones although the TCP does not.  Although these line breaks might provide some slight improvement to the forced alignment we describe below, we do not depend on them.

The three corpora we examined provide further encoding of layout information beyond the top-level regions we focus on in this paper.  All three mark header lines within the running text---often distinguished by larger type and centering---with \texttt{<head>}.  The DTA and WWO record changes of typeface within running text, both at the level of appearance (e.g., roman vs. italic, or Fraktur vs. Antiqua), and at the semantic level (e.g., proper names are often italicized in roman text and in expanded type in Fraktur, but in roman type when surrounded by italics).  The DTA encodes the row and cell structure of some tables but not others. We do not evaluate table layout analysis in this paper due to this inconsistency in the ground truth.

Based on this analysis, we started the process of bootstrapping annotated data for layout analysis with the DTA.  Besides consistently encoding all top-level regions, both its XML transcriptions and page images are available under an open-source license.  We can therefore release the annotations on the DTA produced for this paper as a benchmark dataset.  In addition to experiments on the DTA, we also compiled page-level alignments for a subset of the WWO to test the generalization of models trained on the DTA.  Since the TCP only transcribes a few of the main page regions, we leave further analysis of that corpus for future work.


\section{Annotation by Forced Alignment}
\label{sec:align}

To train and evaluate layout models, we must link digital editions to page images.  This coarse-grained page-level alignment allows us to evaluate models' retrieval accuracy, supporting user queries for images \cite{lee2020newspaper} or footnotes \cite{abuelwafa18:ca}.  Most models and evaluations of layout analysis, however, require a finer-grained assignment of rectangular or polygonal image zones to particular regions.  For both page-level and pixel-level image annotation, we perform forced alignment between the text of digital editions and the output of a baseline OCR system.

For \textbf{page-level} annotation, the DTA already links open-source page images to each page of its 1406 XML editions.  For the WWO, we aligned the ground-truth editions with a corpus of 347,428 OCR'd early modern books from the Internet Archive.  We applied the \texttt{passim} \cite{Smith14:passim} text-reuse analysis system to the ABBYY FineReader transcripts of pages in these books and the 336 XML editions in the WWO.  Processing the pairwise alignments between pages in the IA and in the WWO produced by \texttt{passim}, we selected pairs of scanned and transcribed books such that 80\% of the pages in the scanned book aligned to the XML and 80\% of the pages in the XML aligned with the scanned book.  Furthermore, we required that less than 10\% of the pages in the scanned book align to more than one page in the XML.  This last condition was necessary to exclude editions with pagination differing from that transcribed in the WWO.  In the end, this process produced complete sets of page images for 23 books in the WWO.

Prior to \textbf{pixel-level} image annotation, we have the transcripts of the page regions described above (\S\ref{sec:markup}) but not their locations on physical page image.  We run the Tesseract OCR engine \footnote{\url{https://github.com/tesseract-ocr/tesseract}} on all DTA page images for text line detection and recognition using its publicly available pretrained German model.  The OCR output is then aligned with the ground-truth transcripts from DTA XML in two steps: first, we use \texttt{passim} to perform a line-level alignment of the OCR output with the DTA text.  Next, we perform a character-level forced alignment of the remaining not-yet-aligned OCR output, as well the already aligned text, with the ground-truth text to correct possible line segmentation issues.  In this way, we align regions with one short line---such as page or column number, signature, catchword, and short headings and figure captions---for which \texttt{passim} failed due to limited textual context. This cleanup pass corrected, for example, alignments between a main body region and a note region placed on the left or right.

Once ground-truth transcripts for each text region had been aligned
with the OCR output, region boundaries can be inferred from bounding boxes of the OCR'd text lines.  Assuming that ground-truth transcripts of a region are in reading order, we combined in this order all the bounding boxes and the boundary of the resulting combination is taken as that of the region.

In the digital editions we have examined, figures are not annotated with their exact coordinates or sizes.  Pretrained models such as PubLayNet and Newspaper Navigator can extract figures from page images; however, since they are trained, respectively, on scientific papers and newspapers, which have different layouts from books, the figure detected sometimes also includes parts of other elements such as caption or body near the figure.  To bootstrap annotations for the DTA, we ran Newspaper Navigator on all pages images where the ground truth contained a \texttt{<figure>} element.  Since Newspaper Navigator produces overlapping hypotheses for elements such as figure at decoding time, we check the true number of figures in in the ground truth for the page and then greedily select them in descending order of posterior probability, ignoring any bounding boxes that overlap higher-ranked ones.

The final location accuracy of regions in a page depends on how well Tesseract detected and recognized lines in that page image, how accurate the forced alignment was on noisy OCR output, and how accurately the baseline figure-detection model works.  We therefore manually checked a subset of pages in the DTA for the accuracy of the pixel-level region annotation.  For efficiency, we asked annotators only for binary judgments about the correctness of all regions on a page, rather than asking them to correct bounding boxes or polygons.  We then split the page images into training and test sets (Table~\ref{tab:aligned-regions}).  Since the DTA and Internet Archive images are released under open-source licenses, we  release these annotations publicly.

\begin{table}[htb]
  \centering
  \caption{Pages and regions in the force-aligned, manually checked
    DTA dataset}
  \label{tab:aligned-regions}
  \tabcolsep=5pt
  \begin{tabular}{l|rr}
    Region Type & Train & Test \\
    \midrule
    pages       &  318  &  136 \\
    \midrule
    body        &  340  &  146 \\
    caption      &   33  &   11 \\  
    catchword       &   17  &    4 \\   
    figure     &   53  &   23 \\ 
    note        &  318  &  125 \\
    pageNum     &  313  &  135 \\  
    signature         &   33  &   22 \\    
    title      &  279  &  122 \\
    \bottomrule
  \end{tabular}
\end{table}

\section{Experiments}
\label{sec:experiments}

Having produced alignments between ground-truth editions and page images at the pixel level for the DTA and at the page level for a subset of the WWO, we train and benchmark several page-layout analysis models on different tasks.  First, we consider common pixel-level evaluation metrics.  Then, we evaluate the ability of layout analysis models to retrieve the positions of words in various page regions.  In this way, we aim for a better proxy for end-to-end OCR performance than pixel-level metrics, which might simply capture variation in the parts of regions not overlapping any text.  Then, we evaluate the ability of layout models to retrieve page elements in the full dataset, where pixel-level annotations are not available but the ground-truth provides a set of regions to be detected on each page.  We find a high correlation of these region-level and word-level evaluations with the more common pixel-level metrics. We close by measuring the possibilities for improving accuracy by self-training and the generalization of models trained on the DTA to the WWO corpus.

\subsection{Models}
\label{sec:models}

We trained four models on the training portion of the DTA annotations produced by the forced alignment in \S\ref{sec:align}.  The process produced polygonal boundaries for some regions.  For some experiments, as noted below, we computed the rectangular bounding boxes of these polygons to train on.

Initially, we ran the pretrained PubLayNet \cite{zhong19:icdar} model on the DTA test set, but it failed to find any regions.  We then fine-tuned the \textbf{PubLayNet} F-RCNN weights provided on the DTA training set.  PubLayNet's original training set of over 1 million PDF is only annotated for body, title, list, table, and figures, so it does not produce output for the other region classes.  The best model, using the COCO primary challenge metric mean average precision (mAP=0.824), results from a learning rate of 0.001, batch size of 128, and iteration of 1800.


We trained our own Faster-RCNN (\textbf{F-RCNN}) from scratch on the DTA training set. Our F-RCNN model is based on the ResNet50-FNP-3X baseline provided by Model Zoo \footnote{\url{https://github.com/facebookresearch/detectron2/blob/master/MODEL\_ZOO.md}} and was trained with Detectron2 \cite{wu2019detectron2}. The best performing model has a learning rate of 0.00025, a batch size of 16,  and was trained for 30 epochs.

We also trained two models more directly specialized for page layout analysis: \textbf{Kraken} and \textbf{U-net} (P2PaLA).  We adopted both systems' default architecture definitions and training hyperparameters.  Page images were binarized and scaled to a height of 1200 and 1024 pixels for Kraken and U-net, respectively.  Both models were trained with binary cross-entropy loss and the Adam optimizer, with learning rate $20^{-5}$ for 50 epochs with Kraken, and $10^{-4}$ for 200 epochs with U-net.  To allow the models to generalize better on unseen samples, data augmentation was used by applying on-the-fly random transformations on each training image.
%

\subsection{Pixel-level Evaluations}
\label{sec:pixel-eval}

To investigate whether regions annotated with polygonal coordinates
have some advantage over annotation with rectangular coordinates, we
trained the Kraken and U-net models on both annotation types. (The F-RCNN models only infer rectangles.) These models were
trained and evaluated on the data defined in
Table~\ref{tab:aligned-regions}.
Table~\ref{tab:pixel-evaluation} reports figures for standard
region segmentation metrics\,\cite{long2015fully}: pixel
accuracy (p\_acc), mean pixel accuracy (m\_acc), mean Jaccard Index (m\_iu), and
frequency-weighted Jaccard Index (f\_iu), for evaluating layout models for
systems trained on different annotation types.

\begin{table}[htb]
  \centering
  \caption{Evaluation on four pixel-level metrics: PubLayNet fine-tuned and F-RCNN, Kraken, and U-net trained on DTA data.  The first two models require rectangular bounding boxes at training time; the latter two may use polygons or rectangles.}
  \label{tab:pixel-evaluation}
  \tabcolsep=5pt
  \begin{tabular}[t]{lrrrrrr}
    \toprule
    & PubLayNet & F-RCNN & \multicolumn{2}{c}{Kraken} & \multicolumn{2}{c}{U-net} \\
           & Rect & Rect & Poly  &  Rect &  Poly &  Rect \\
    \midrule
    p\_acc & 0.966  & 0.975 & 0.909 & 0.938 & 0.960 & 0.960 \\
    m\_acc & 0.973  & 0.894 & 0.511 & 0.537 & 0.928 & 0.946 \\
    m\_iu  & 0.890  & 0.781 & 0.480 & 0.516 & 0.810 & 0.790 \\
    f\_iu  & 0.886  & 0.881 & 0.858 & 0.907 & 0.932 & 0.933 \\
    \bottomrule
  \end{tabular}%
  \hfill
\end{table}

\begin{table}[htb]
  \centering
  \caption{Pixel-level evaluation by region type: PubLayNet fine-tuned and F-RCNN, Kraken, and U-net trained on DTA data.  PubLayNet does not output region types not in its original training data; Kraken produces no output for the smaller regions.}
  \label{tab:pixel-regions}
  \tabcolsep=5pt
  \begin{tabular}[t]{l|rr|rr|rr|rr}
    \toprule
    & \multicolumn{2}{c}{PubLayNet} & \multicolumn{2}{c}{F-RCNN} & \multicolumn{2}{c}{Kraken} & \multicolumn{2}{c}{U-net} \\
    Region & p\_acc & iu & p\_acc & iu & p\_acc & iu & p\_acc & iu \\
    \midrule
    body    & 0.96 & 0.92 & 0.97 & 0.94 & 0.93 & 0.91 & 0.96 & 0.94 \\
    caption  & ---  & ---  & 0.91 & 0.77 & --- & --- & 0.78 & 0.60 \\
    catchword   & ---  & ---  & 0.50 & 0.40 & --- & --- & 0.51 & 0.33 \\
    figure & 0.99 & 0.90 & 0.98 & 0.89 & --- & --- & 0.94 & 0.74 \\
    note    & ---  & ---  & 0.98 & 0.92 & 0.82 & 0.77 & 0.93 & 0.88 \\
    pageNum & ---  & ---  & 1.00 & 0.86 & --- & --- & 0.94 & 0.68 \\
    signature     & ---  & ---  & 0.82 & 0.60 & --- & --- & 0.37 & 0.26 \\
    title  & 0.97 & 0.84 & 0.99 & 0.87 & --- & --- & 0.96 & 0.74 \\
    \bottomrule
  \end{tabular}%
\end{table}

From the results of Table\,\ref{tab:pixel-evaluation}, we can
see there is not a significant difference between using rectangular or
polygonal annotation for regions, but there is a substantial
difference between the performance of the systems. Not shown in the table is the out-of-the-box PubLayNet, which is not able to detect any content in the dataset, but its performance improved dramatically after fine-tuning. Our own F-RCNN provides comparable results for the regions detectable in the fine-tuned PubLayNet, while it also detects 5 other regions. The differences among systems are
more evident in
Table\,\ref{tab:pixel-regions}, where Kraken's
predictions detected only ``body'' and ``note'' and failed for the
remaining (small) regions and the fine-tuned PubLayNet likewise predicted only a subset of the page regions. For this reason, we consider only the F-RCNN and U-net models in later experiments.

\begin{table}[htb]
  \centering
  \caption{Using F-RCNN on our annotated test set. AP @ iu [0.5:0.95].}
  \label{tab:OurFRCNNResult}
  \tabcolsep=5pt
  \begin{tabular}{l|rrrr}
    \toprule
    & \multicolumn{4}{c}{F-RCNN} \\
     Region Type & AP  \\
    \midrule
    body         &   0.888  \\
    caption  &  0.638 \\
    catchword        &  0.523  \\
    figure      & 0.788  \\
    note        &  0.868  \\
    pageNum      &   0.829  \\
    signature          &  0.454  \\
    title      &  0.792  \\
    OVERALL(mAP)     & 0.723    \\
    \bottomrule
  \end{tabular}%
 
\end{table}


\subsection{Word-level Evaluations}

While pixel-level evaluations focus on the layout analysis task, it is
also worthwhile to measure a proxy for end-to-end OCR performance.

Using the positions of word tokens in the DTA test set as detected by
Tesseract, we evaluate the performance of regions predicted by the
U-net model considering how many words of the reference region fall
inside or outside the boundary of the predicted
region. Table\,\ref{tab:WrdLevUnetResults} shows word-level retrieval
results in terms of recall (Rc), precision (Pr) and F-measure (F1)
metrics for each region type.

\begin{table}[!htb]
  \centering
  \caption{Word-level retrieval results for the different region types
    predicted by the U-net model.}
  \label{tab:WrdLevUnetResults}
  \tabcolsep=5pt
  \begin{tabular}{l|rrr}
    \toprule
    Region Type & Rc & Pr & F1 \\
    \midrule
    body    & 0.91 & 0.98 & 0.94 \\
    caption & 0.63 & 0.73 & 0.66 \\
    catchword   & 0.33 & 0.33 & 0.33 \\
    note    & 0.85 & 0.98 & 0.91 \\
    pageNum & 0.81 & 0.81 & 0.81 \\
    signature     & 0.26 & 0.44 & 0.31 \\
    title   & 0.92 & 0.97 & 0.94 \\
    \bottomrule
  \end{tabular}
\end{table}

\subsection{Region-level Evaluations}

This is a simpler evaluation since it does not require word-position
coordinates as the word-level case, considering only for each page
whether its predicted region types are or not in the page
ground-truth.
Therefore, we can use the already trained layout models for inferring
the regions on the entire DTA collection (composed of 500K page
images) and also on the out-of-sample WWO dataset containing more than
5,000 pages with region types analogous to DTA.


Since PubLayNet and Kraken do not detect all the categories we want to
evaluate, we perform this region-level evaluation using only the
U-net and F-RCNN models, which were already trained on the 318
annotated pages of the DTA collection.
%
To evaluate the performance over the entire DTA dataset and on WWO
data, we use region-level precision, recall, and F1 metrics.
Table.\,\ref{tab:RegLvResDTA-WWO} reports these evaluation metrics
for the regions detected by these two models on the entire DTA and WWO
datasets.

\begin{table}[htb]
  \centering
  \caption{Region-level retrieval results (Pr, Rc and F1) on the
    entire DTA collection and WWO data using U-net and F-RCNN. IoU
    threshold detection meta-parameter of F-RCNN model was set up to
    0.9 and 0.5 for DTA and WWO respectively.  The WWO does not annotate running titles or column heads, and the WWO test books contain figures but no captions.}
  \label{tab:RegLvResDTA-WWO}
  \tabcolsep=4pt
  \begin{tabular}{l|rrr|rrr}
    \toprule
    \multirow{3}{*}{Reg. Type}
                & \multicolumn{6}{c}{U-net} \\
                & \multicolumn{3}{c}{DTA} & \multicolumn{3}{c}{WWO} \\
            &   Rc &   Pr &   F1 &   Rc &   Pr &   F1 \\
    \midrule
    body    & 0.89 & 0.89 & 0.89 & 0.95 & 0.95 & 0.95 \\
    caption & 0.85 & 0.37 & 0.52 &  --- &  --- &  --- \\ 
    catchword   & 0.89 & 0.63 & 0.74 & 0.77 & 0.52 & 0.62 \\
    colNum  & 0.20 & 0.25 & 0.22 &  --- &  --- &  --- \\ 
    figure  & 0.77 & 0.81 & 0.79 & 0.53 & 0.43 & 0.48 \\
    note    & 0.97 & 0.97 & 0.97 & 0.84 & 0.84 & 0.84 \\
    pageNum & 0.96 & 0.68 & 0.80 & 0.98 & 0.45 & 0.61 \\
    signature & 0.65 & 0.48 & 0.56 & 0.32 & 0.25 & 0.28 \\
    title   & 0.97 & 0.69 & 0.81 &  --- &  --- &  --- \\ 
    \bottomrule
  \end{tabular}%
  \begin{tabular}{|rrr|rrr}
    \toprule
    \multicolumn{6}{|c}{F-RCNN} \\
    \multicolumn{3}{|c}{DTA} & \multicolumn{3}{c}{WWO} \\
      Rc &   Pr &   F1 &   Rc &   Pr &   F1 \\
    \midrule
    0.92 & 1.00 & 0.96 & 0.99 & 1.00 & 0.99 \\
    0.06 & 0.88 & 0.80 &  --- &  --- &  --- \\ 
    0.26 & 0.96 & 0.41 & 0.54 & 0.48 & 0.52 \\
    0.81 & 0.84 & 0.82 &  --- &  --- &  --- \\
    0.59 & 0.32 & 0.41 & 0.50 & 0.00 & 0.00 \\
    0.61 & 0.64 & 0.62 & 0.42 & 0.18 & 0.26 \\
    0.81 & 0.79 & 0.80 & 0.61 & 0.23 & 0.34 \\
    0.12 & 0.85 & 0.21 & 0.13 & 0.14 & 0.13 \\
    0.52 & 0.78 & 0.54 &  --- & ---  &  --- \\
    \bottomrule
  \end{tabular}
\end{table}


The F-RCNN model can find all the graphic figures in the ground truth;
however, since it also has a high false positive value, the precision
for \texttt{figure} is 0 at confidence threshold of 0.5. In general,
as can be observed in Table~\ref{tab:RegLvResDTA-WWO}, F-RCNN seems to
generalize less well
than U-net
on several region types in both the DTA and WWO.

\subsection{Improving accuracy with self-training}

While the amount of data we can manually label at the pixel level is small, the availability of page-level information on regions in the whole corpus allows us to improve these models by self-training.  Instead of simply adding in potentially noisy automatically labeled images to the training set, we can restrict the new training examples to those pages where all regions have been successfully detected.  In analyzing one iteration of this procedure, we find that overall pixel-level metrics improve slightly, but improve substantially for particular regions (Table~\ref{tab:globPxLvTrIt}).

\begin{table}[!htb]
  \centering
  \caption{Pixel-level evaluation results for U-net with two
    self-training rounds on the 136 annotated pages of the DTA test set:
    globally (left) and by region (right).}
  \label{tab:globPxLvTrIt}
  \tabcolsep=5pt
  \begin{tabular}[t]{lrr}
    \toprule
    Metric &   Round 1 &  Round 2 \\
    \midrule
    p\_acc &     0.960 &    0.964 \\
    m\_acc &     0.928 &    0.934 \\
    m\_iu  &     0.810 &    0.845 \\
    f\_iu  &     0.932 &    0.937 \\
    \bottomrule
  \end{tabular}%
  \hspace{20pt}%
  \begin{tabular}[t]{l|rr|rr}
    \toprule
    & \multicolumn{2}{c|}{Round 1} & \multicolumn{2}{c}{Round 2} \\
    Region type & p\_acc & iu & p\_acc & iu \\
    \midrule
    body    & 0.960 & 0.940 & 0.968 & 0.952 \\
    caption  & 0.783 & 0.596 & 0.704 & 0.548 \\
    catchword   & 0.513 & 0.334 & 0.657 & 0.447 \\
    figure & 0.937 & 0.735 & 0.966 & 0.701 \\
    note    & 0.928 & 0.880 & 0.942 & 0.902 \\
    pageNum & 0.937 & 0.683 & 0.960 & 0.740 \\
    signature     & 0.369 & 0.262 & 0.481 & 0.410 \\
    title  & 0.961 & 0.735 & 0.920 & 0.827 \\
    \bottomrule
  \end{tabular}
\end{table}

Likewise, we see similar improvements in many region types in the word- and region-level evaluations (Tables~\ref{tab:WrdLevUnetResultsTrItr} and \ref{tab:RegLvUnetResDTATrItr}).  
Notably, the accuracy of detecting figures declines with self-training, which we find is due to only two images with figures appearing in the set of pages with all regions correctly detected.  Balancing different page layouts during self-training might mitigate this problem.

\begin{table}[!htb]
  \centering
  \caption{Word-level retrieval results for U-net with two
    self-training rounds on the 136 annotated pages of the DTA test set.}
  \label{tab:WrdLevUnetResultsTrItr}
  \tabcolsep=5pt
  \begin{tabular}{l|rrr|}
    \toprule
    & \multicolumn{3}{c|}{Round 1} \\
    Region Type &   Rc &   Pr &   F1 \\
    \midrule
    body        & 0.91 & 0.98 & 0.94 \\
    catchword       & 0.33 & 0.33 & 0.33 \\
    figure      & 0.63 & 0.73 & 0.66 \\
    note        & 0.85 & 0.98 & 0.91 \\
    pageNum     & 0.81 & 0.81 & 0.81 \\
    signature         & 0.26 & 0.44 & 0.31 \\
    title      & 0.92 & 0.97 & 0.94 \\
    \bottomrule
  \end{tabular}%
  \begin{tabular}{rrr}
    \toprule
    \multicolumn{3}{c}{Round 2} \\
    Rc & Pr & F1 \\
    \midrule
    0.94 & 1.00 & 0.97 \\
    0.83 & 1.00 & 0.89 \\
    0.52 & 0.53 & 0.50 \\
    0.89 & 0.95 & 0.92 \\
    0.87 & 0.87 & 0.87 \\
    0.41 & 0.56 & 0.44 \\
    0.88 & 0.98 & 0.91 \\
    \bottomrule
  \end{tabular}%
\end{table}

\begin{table}[!htb]
  \centering
  \caption{Region-level retrieval results for U-net with self-training
    on the 136 annotated pages of DTA's test set.}
  \label{tab:RegLvUnetResDTATrItr}
  \tabcolsep=5pt
  \begin{tabular}{l|rrr}
    \toprule
    & \multicolumn{3}{c}{Round 1} \\
    Region Type &   Rc &   Pr &   F1 \\
    \midrule
    body    & 0.99 & 0.99 & 0.99 \\ 
    caption  & 1.00 & 0.82 & 0.90 \\ 
    catchword   & 1.00 & 1.00 & 1.00 \\ 
    figure & 0.86 & 0.86 & 0.86 \\ 
    note    & 1.00 & 1.00 & 1.00 \\ 
    pageNum & 1.00 & 0.92 & 0.96 \\ 
    signature     & 0.91 & 0.57 & 0.70 \\ 
    title  & 1.00 & 0.92 & 0.96 \\ 
    \bottomrule
  \end{tabular}%
  \begin{tabular}{|rrr}
    \toprule
    \multicolumn{3}{|c}{Round 2} \\
    Rc &   Pr &   F1 \\
    \midrule
              1.00 & 1.00 & 1.00 \\ 
               1.00 & 0.62 & 0.77 \\ 
               1.00 & 1.00 & 1.00 \\ 
               0.82 & 0.82 & 0.82 \\ 
              0.96 & 0.96 & 0.96 \\ 
               0.99 & 0.95 & 0.97 \\ 
               0.95 & 0.72 & 0.82 \\ 
               1.00 & 0.96 & 0.98 \\ 
    \bottomrule
  \end{tabular}
\end{table}

\subsection{Correlation of word/region-level and pixel-level metrics}

To analyze the correlation of word/region- and pixel-level metrics,
the Pearson correlation coefficient has been computed for each region
between word/region-level F-score (F1) and pixel Jaccard index (iu)
values obtained on the test set of 136 annotated pages (see
Table\,\ref{tab:aligned-regions}). The reported linear correlation
coefficients in Table\,\ref{tab:Coor_PxAccVsPRWL} in general show
significant correlations (with p-values lower than $10^{-10}$) between
these different metrics.

\begin{table}[!htb]
  \centering
  \tabcolsep=5pt
  \caption{Pearson correlation coefficients between the Jaccard-Index
    pixel-level metric (iu), and word- \& region-level F-score (F1)
    metrics. This study was conducted on DTA test set with the
    U-net model.}
  \label{tab:Coor_PxAccVsPRWL}
  \vspace{-.8em}
  \begin{tabular}[t]{l|cr}
    \toprule
    Region Type & Word Level \\
    \midrule
    body    & 0.980 \\
    note    & 0.970 \\
    title  & 0.812 \\
    pageNum & 0.316 \\
    signature     & 0.524 \\
    figure  & 0.962 \\
    catchword   & 0.792 \\
    \bottomrule
  \end{tabular}%
  \begin{tabular}[t]{|cr}
    \toprule
    Region Level \\
    \midrule
              0.906 \\
              0.903 \\
              0.861 \\
    
              0.569 \\
              0.702 \\
              0.966 \\
              0.916 \\
    \bottomrule
  \end{tabular}
\end{table}
    

\begin{figure}[!htb]
  \centering
  \includegraphics[width=.493\textwidth]{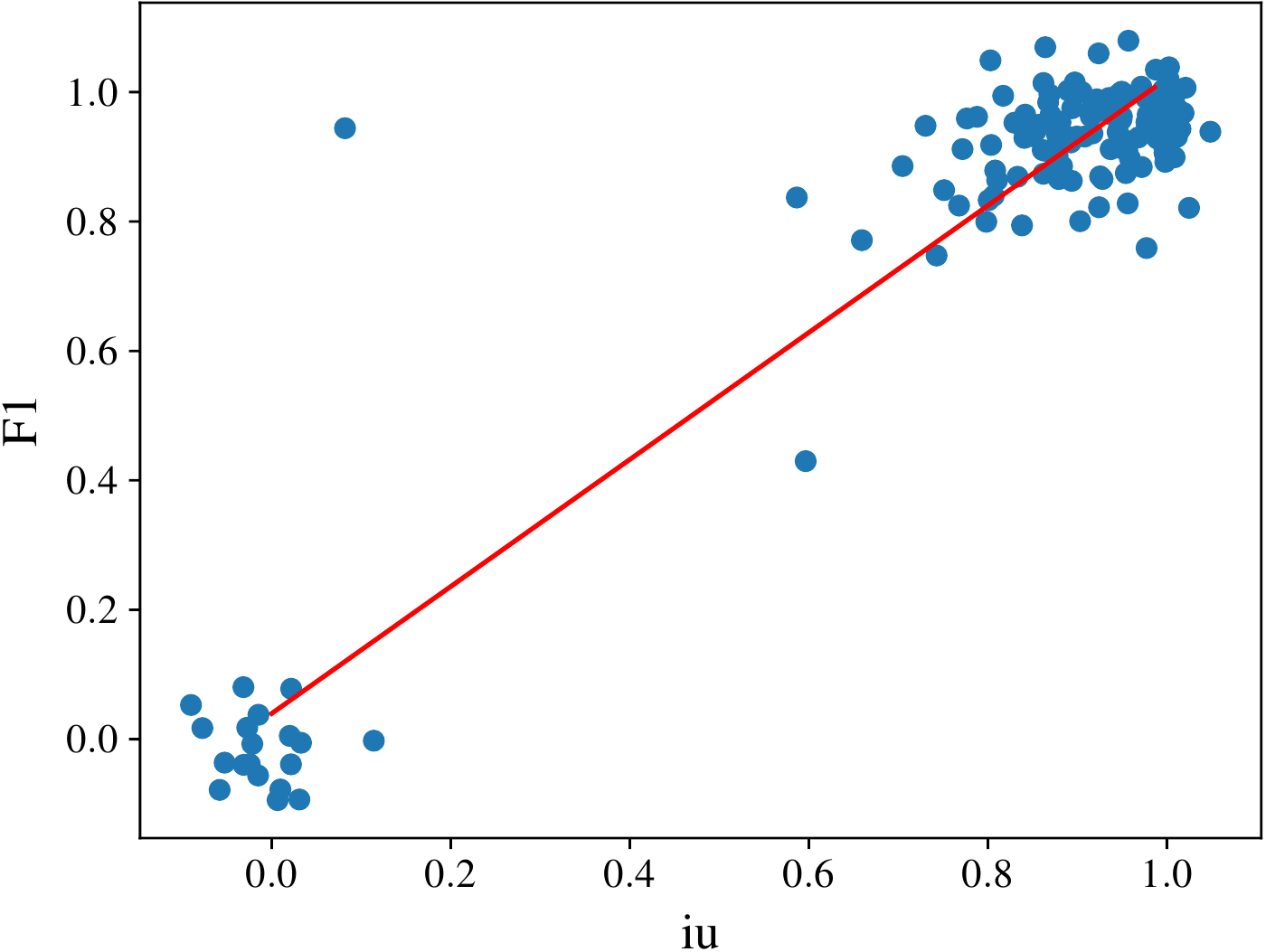}%
  \hfill%
  \includegraphics[width=.49\textwidth]{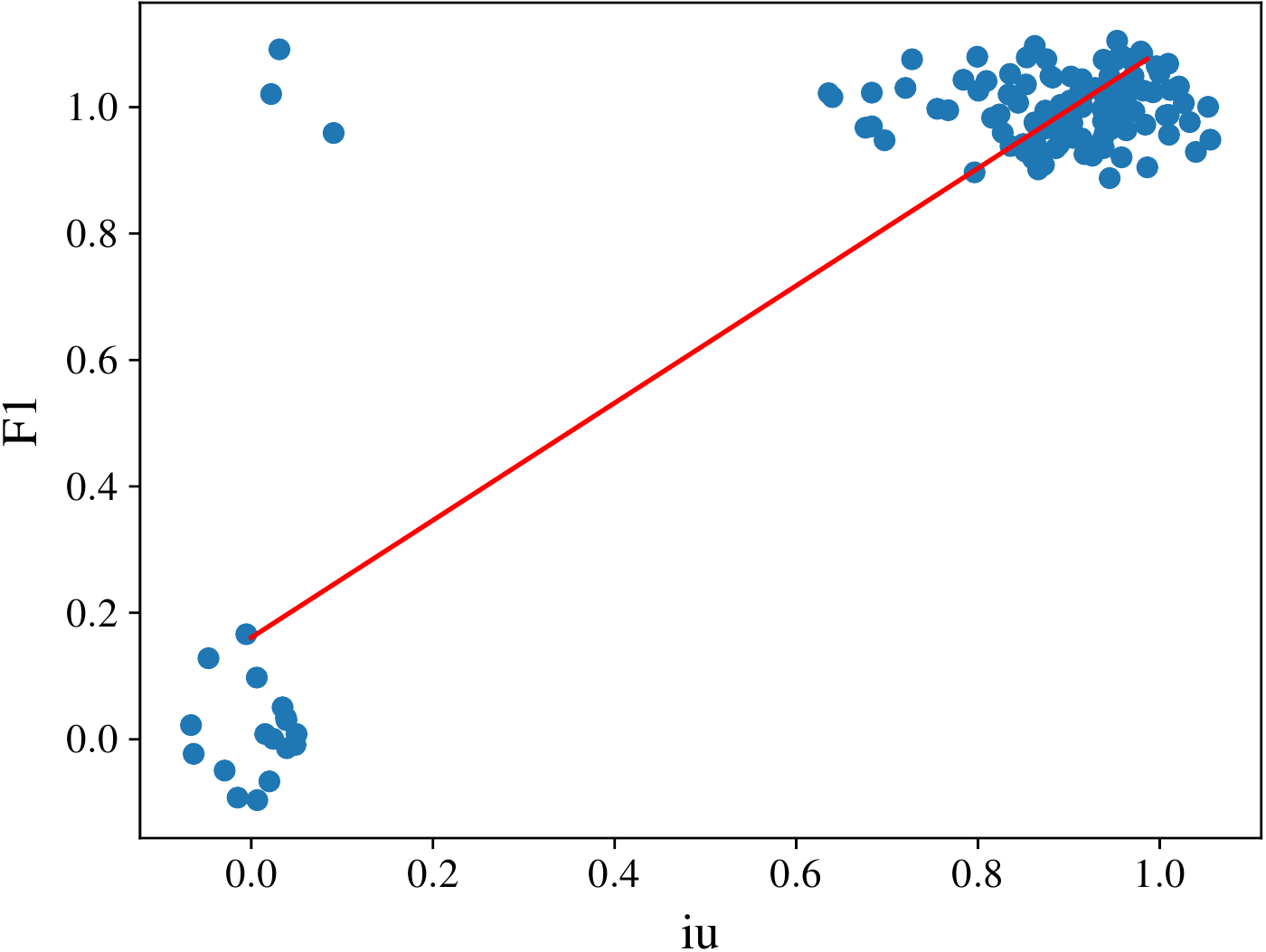}%
  \caption{Scatter plots with linear correlation of Jaccard Index (iu)
    versus word-level (left) and region-level F-score (right) for
    region type ``note''. These were produced using U-net on DTA's
    test set data.}
  \label{fig:corrcoeff-plots}
\end{figure}


\section{Conclusions}

We found that several broad-coverage collections of digital editions can be aligned to page images in order to construct large testbeds for document layout analysis.  We manually checked a sample of regions annotated at the pixel level by forced alignment.  We benchmarked several state-of-the-art methods and showed a high correlation of standard pixel-level evaluations with word- and region-level evaluations applicable to the full corpus of a half million images from the DTA.  We publicly released the annotations on these open-source images at \url{https://github.com/NULabTMN/PrintedBookLayout}. Future work on these corpora could focus on standardizing table layout annotations; on annotating sub-regions, such as section headers, poetry, quotations, and contrasting typefaces; and on developing improved layout analysis models for early modern print.


%

\bibliographystyle{splncs04}
\bibliography{bootstrap-layout}

\end{document}